\documentclass[conference]{IEEEtran}
\IEEEoverridecommandlockouts
\usepackage{cite}
\usepackage{amsmath,amssymb,amsfonts}
\usepackage{algorithmic}
\usepackage{graphicx}
\usepackage{textcomp}
\usepackage{xcolor}
\usepackage{epstopdf}
\usepackage{lipsum}
\usepackage{url}
\def\BibTeX{{\rm B\kern-.05em{\sc i\kern-.025em b}\kern-.08em
    T\kern-.1667em\lower.7ex\hbox{E}\kern-.125emX}}
\begin{document}

\title{NAttack! Adversarial Attacks to bypass a GAN based classifier trained to detect Network intrusion
\thanks{This work supported in part by a gift from IBM Research}
}

\author{\IEEEauthorblockN{Aritran Piplai, Sai Sree Laya Chukkapalli, Anupam Joshi}
\IEEEauthorblockA{\textit{Department of Computer Science} \\
\textit{University of Maryland Baltimore County}\\
Baltimore, USA \\
apiplai1@umbc.edu, saisree1@umbc.edu, joshi@umbc.edu}

}

\maketitle

\begin{abstract}
With the recent developments in artificial intelligence and machine learning, anomalies in network traffic can be detected using machine learning approaches. Before the rise of machine learning,network anomalies which could imply an attack, were detected  using  well-crafted  rules.  An  attacker who has knowledge in the field of cyber-defence, could make educated guesses  to  sometimes  accurately predict which particular features of network traffic data the cyber-defence mechanism is looking at. With this information, the attacker can circumvent a rule-based cyber-defense system. However, after the advancements of machine learning for network anomaly, it is not easy for a human to understand how to bypass a cyber-defence system. Recently, adversarial attacks have become increasingly common to defeat machine learning algorithms. In this paper, we show that even if we build a classifier and train it with adversarial examples for network data, we can use adversarial attacks and successfully break the system. We  propose  a  Generative  Adversarial  Network(GAN)based algorithm to generate data to train an efficient neural network based classifier, and we subsequently break the system using adversarial attacks.
\end{abstract}


\section{Introduction}
Network intrusions have resulted in large-scale damages in major organizations over a significant period of time. Cisomag \cite{cisomag}, in their report, states that a survey taken by 5400 organizations in US and Europe says that 61\% of the companies have faced a cyber-intrusions. The surge in the number of attacks has forced the financial losses endured by organizations due to cyber-attacks, to rise five fold. In 2017, an attack on the US Electricity Grid was performed using multi-stage intrusion by threat actors. \cite{cyber_mdx}. This led the attackers to gain access to the control switches with the help of which they could have disrupted the power-flow of the entire nation. As network intrusions have become common, the intrusion detection systems have also evolved. They are increasingly using Machine Learning based approaches to detect attacks. A report by BDTechtalks \cite{bdtechtalks} describes how email phishing defence systems are finding it difficult to prevent malicious emails from being communicated, and are having to rely on AI. Attackers of this kind are increasingly using `Distributed Spam Distraction' and `polymorphic attacks'. This is making the job of human operators very difficult as `spam distraction' is stressing out the defense systems, and `polymorphic attacks' are being used to `cloak the identity' of the malwares. AI solutions are dealing with these problems more effectively than human crafted rules. Security Intelligence \cite{security_intelligence} states multiple reasons why AI might be paramount for cybersecurity. One of the main reasons being, malware is becoming increasingly complex. This is making the job of cybersecurity engineers increasingly difficult, to create rules which would help identify an attack. Another reason why AI will become important, is that effective reporting of cyber-incidents has resulted in a large quantity of data which could be used to train AI algorithms. For example, MIT Lincoln Labs has been releasing Intrusion-Detection datasets since 1998. \cite{Lincolnlabs}

With AI being used in cyber-defense systems, it is also being increasingly used for offense. There are reports about how attackers have used GANs to mimic the voice of a trusted person and dupe people.\cite{security_boulevard}. This, along with other similar attacks, have lead security engineers to give adversarial attacks increased attention. Israel's cyber-chief, at cybertech conference, stated that adversarial AI has attempted to hack automotives to make them `behave in a manner contrary to their programming'. \cite{jposyt}. 

In this paper, we show that adversarial attacks can be used to defeat machine learning based intrusion detection systems.  Specifically, we describe how a classifier yielding very good performance in detecting a network intrusion alert, fails to do so when faced with adversarial attacks.We focus on a dataset of a network intrusion alert system.\cite{knowledge_pit} The dataset has features related to a cyber-event. The first stage of our experiment involves building a robust neural network classifier with the help of GANs \cite{GAN}. GANs involve two neural networks competing against each other, where one, the generator, is trying to fool the other, the discriminator, with generated adversarial examples. The discriminator, in turn, aims to not get fooled by the generated adversarial examples. We use GANs to generate adversarial examples during the training phase, hoping to make the classifier aware of those adversarial examples. GANs also help us in terms of dealing with the class imbalance problem,\cite{class_imbalance} which is very common for training a dataset of network intrusion alerts, thereby increasing classification scores for both the classes. In the second part of our experiment we describe how we can use `Fast Gradient Sign Method' \cite{Goodfellow_fast_gradient} to perform adversarial attacks on a classifier with high accuracy. This helps in masquerading the cyber-event samples by perturbing the data slightly, so that to the neural network based classifier it seems that the event is a `non-attack' sample, when in reality it is an `attack' sample. The key contributions of our research are that this paper states that using GANs we can create classifiers which yield very high scores when used to detect network intrusion. The other contribution of this paper, is that we show that even such a classifier is vulnerable to carefully crafted adversarial attacks. 

We organize the paper as follows. Section \ref{RelatedWork} talks about the works of other researchers in this domain. Section \ref{Methodology} talks about the architecture of our experiment, the intuition, and the mathematical foundation of the system. Section \ref{Experimental Setup} discusses our findings from this experiment. Section \ref{Conclusion} talks about the scope of future extensions of this research.

\section{Related Work}
\label{RelatedWork}
In the recent times network attacks have become a major challenge in the field of security. A lot of effort is being put by researchers to mitigate network based attacks. One such solution is the intrusion detection system (IDS) which alerts the user whenever an abnormal event happens. The IDS is said to be more robust and secure if it detects the attack quickly and also learns new attacks. There are two ways of detecting attacks such as rule-based detection and the other is machine learning methods. The rule-based detection basically identifies known type of attacks. The most commercially used IDS is Snort\cite{Snort} which detects an attack based on content matching the rules from the file that contains characteristics of each type of an attack. The main drawback for this method is that it cannot identify new types of attacks.

In order to overcome this drawback machine learning methods are widely being used for identifying unknown type of attacks. Lee and Stoflo \cite {Lee_Stoflo} published a framework to collect features for network intrusion detection, which could be used by machine learning algorithms. The machine learning methods that are used for attack detection  can be categorized as supervised learning and unsupervised learning. The supervised learning algorithm collects the data from network and classifies them as an abnormal event or normal event based on the  behavior. There are various types of supervised learning algorithms such as Naïve Bayes\cite{naive_bayes}, Random Forest\cite{random_forest},Support Vector Machine (SVM) \cite{Jha_Ragha}, Decision Tree, etc that are used for attack detection. A performance evaluation is done for the above classification algorithms by Belouch et al. \cite{Belouch} to compare the accuracy for detection, prediction time and it is observed that Random Forest showed better performance in terms of accuracy. The only drawback in supervised learning algorithm is that they require labelled data for training the algorithm. In order to overcome this issue we use unsupervised methods for detecting the anomalies in the network.

The traditional unsupervised learning methods draw inferences from the unlabeled data for training the algorithm. Most of the unsupervised methods are based on clustering and outliers detection. In clustering the most popular approach in unsupervised learning is K-means where the data is divided into clusters based on certain similarities. Here k is defined as number of clusters. In the paper \cite{kmeans} the authors used the K-means approach to detect the abnormal behaviour from the KDD99 dataset based on Euclidean distance in order to control fraud in the network traffic. For the outlier detection the most commonly used approach to leverage anomaly detection is Principal Component Analysis(PCA) \cite{PCA}. This is a multivariate method where dimensionality reduction is done to lower the dimensional space while preserving the information from the data.

The other set of unsupervised learning methods that gained lot of popularity for their performance are deep-learning based as they require only raw data for training. For example, Restricted Boltzmann machine \cite{RBM} was used as an intrusion detection system in smart cities to have better accuracy results in attack detection by optimizing the hyperparameters. Similarily, AutoEncoders\cite{AE} is another popular deep learning method being used since it provides better classification results when compared to the traditional deep learning methods.

As applications of AI have become popular, research has also been done on how AI applications can be susceptible to adversarial attacks. Xu et. al \cite {Xu} have discussed in their paper, the adversarial attacks that can affect neural network based classifiers in the domain of images and graphs. In the domain of images, significant work has been done related to adversarial attacks as can be exemplified by Eykholt et. al \cite{Eykholt}. Research on adversarial attacks has also been done for sequence to sequence models. He and Glass state in their paper,\cite{He_Glass} that they have made a model that calculates input sequences that will generate incredible text sequences. However, images are different from network intrusion data. For instance, an image has pixels, each of them is equally important, and any of them can be changed to a certain extent without significant change occurring to the appearance of the entire image. However, a cyber-event data has specific features, some of them may be sensitive, and changing them even slightly may make the change detectable.

\section{Methodology}
\label{Methodology}
\begin{figure*}
    \centering
    \includegraphics[width=0.6\textwidth, scale=10]{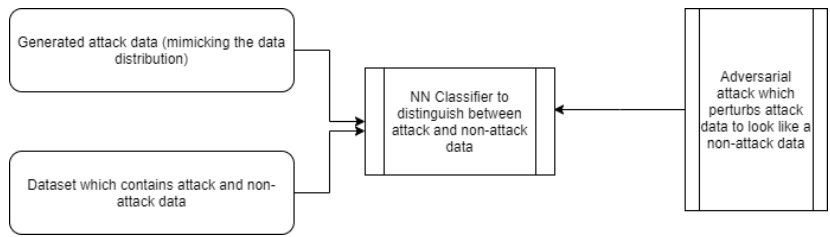}
    \caption{Basic Architecture of our experiment}
    \label{neuralarch}
\end{figure*}

In this section we describe the neural model we use to distinguish between an actual attack and a non-attack scenario. We then describe our findings related to how difficult or easy it is for an attacker to break the machine learning system using simple adversarial attacks. Figure \ref{neuralarch} describes the high-level architecture of our experimental setup. We first train a classifier using GAN, which distinguishes between `attack' and a `non-attack'. Then, we using `Fast Sign Gradient Method' \cite{Goodfellow_fast_gradient} to effectively perturb the `attack' samples so that the classifier trained in the previous step, is forced to classify them as `non-attack'. 
\subsection{Dataset Description and Pre-processing}
We have used the data from the IEEE BigData 2019 Cup:Suspicious Network Event Recognition challenge. \cite{knowledge_pit} The data set given was built based  on the data captured from the network traffic events logged by the security systems and was sponsored by Security on-Demand and QED Software. Initially,the compressed data set comprised of training data, test data and localized alerts present in the auxillary table. It was restored based on the `alert\_id' column that is common for all three files by adding the columns of localized alert file to the training and test data files when there is match based on the identifiers present in the `alert\_id' column. Some of the important columns in the dataset are described here for better understanding like the binary column named `notified' is present only in the training data states whether the client was notified about the alert or not. The severity of the event is categorized and specified in the column `categoryname'. Data preprocessing techniques like filling the missing values in the 27 columns of the training data and 26 columns of test data after being merged with localized alerts based on `alert\_id' have been applied. Then normalization technique such as rescaling is used to get the data into a common scale as the columns in the data set initially had values in different ranges.
\label{dataset_descr}
\subsection{Neural Network Classifier}
The dataset, as described in Section \ref{dataset_descr},has multiple features which describe each sample point. The field `notified' is a binary field where `0' represents that the data point is a non-attack, and `1' represents an actual attack. There is also an auxillary dataset, called `localized alerts', as mentioned in Section  \ref{dataset_descr} which has a one-to-one mapping with the `alert-id' feature of the primitive dataset. The primary dataset has 6 fields, while the localized alert dataset has 20 features. Each data point in the primary dataset, upon being merged with the localized alert data corresponding with the `alert-id' feature of the datapoint, has 26 features. This can be represented as a 26 dimensional vector, and neural networks can be used to classify these vectors into `true attack' and `non-attack'. 
\subsubsection{GAN to boost classifier performance}
\label{GAN_purpose_1}
The merged dataset had a class imbalance problem. The samples for `true attack' were significantly lower in number than the samples for `non-attack'.  The purpose of using GAN for this problem is two-fold. As stated in \cite{class_imbalance}, GANs can be effectively used to solve a class imbalance problem inherent in the dataset. Any neural network based classifier is susceptible to bias introduced due to class imbalance. GANs are used to generate training samples for the minority class, to make the classifier robust. The minority class in this case, is the class of the true attack samples. Given a basic feed-forward neural network, the classifier would have been biased towards the majority class, i.e, the class of `non-attack' samples. During our experiments we found out that the F-1 scores for attack samples were significantly lower, which means given any network event, the classifier would be more likely to classify it as an `non-attack' sample. The adversarial examples generated for the attack class during training time, forces the neural network to distinguish between attack and non-attack samples in a more robust way.
\subsubsection{GAN to help protect against Adversarial Attack}
\label{GAN_purpose_2}
\begin{figure*}
    \centering
    \includegraphics[width=0.5\textwidth]{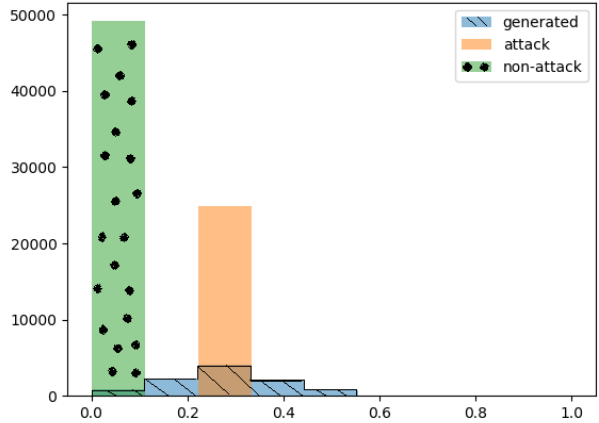}
    \caption{Example of data distribution for a particular feature where the class boundaries are well defined}
    \label{data_distr}
\end{figure*}

The second reason why we use GAN to train our classifier, is because it trains the neural network with adversarial examples. In the later part of the paper, we want to break the AI classifier. To make the claim that the attack is dangerous and effective, we first have to make sure that the classifier has been trained so that simple adversarial attacks do not make an impact. When GANs generate adversarial examples, it tries to mimic the distribution of the data, and tries to generate from that distribution. Simple adversarial attacks forces a classifier to make the mistake of mispredicting the output class of a sample by exploiting class boundaries. With our training method, generated samples help to relax those class boundaries so that the adversarial attacks become difficult. For example in Figure \ref{data_distr}, we can see that the class boundaries are rigid for those samples and that particular feature. Hypothetically, if we were to perform an adversarial attack based on this data, we could have perturbed the value to 0.199 and it would not have been classified in the attack class. The generated samples for the attack class, helps to relax those boundaries, as we have some samples for the attack class now, where the values do not lie between 0.2 and 0.4. 
\subsection{Neural Network Architecture}
The basic GAN has two components. One is a generator and the other is a discriminator. We have described informally the functionality and the purpose of the generator in Sections \ref{GAN_purpose_1} and \ref{GAN_purpose_2}. The other component, the discriminator, helps learns the original distribution, and also learns how to identify the generated samples as attacks. This makes the discriminator a very well-informed classifier, as it sees the data from the original distribution, and also the generated data. We use this discriminator as our classifier, and we describe how we subsequently break this classifier in the following sections. In this section we describe formally the discriminator and the generator. 
A conventional GAN has two neural networks, a discriminator and a generator. The generator tries to generate fake data, while the discriminator tries to model how to differentiate between the real data and the fake data. In our case, the generator generates samples mimicking the data distribution. These samples are the adversaries which the discriminator wants to classify as `attack'. The aim of the generator is to generate `attack' samples, which are the adversaries, such that the discriminator confuses them to be `non-attack' data points. The aim of the discriminator is to classify the generated adversaries as `attack' samples. 
In conventional GAN, there are two loss metrics. 
\begin{itemize}
     \item Generator loss: Generator loss is calculated only for the generated data, as opposed to both real and generated data for the discriminator loss. The generated data is given the label 1, and the output of the discriminator is calculated when the generated data is served as an input. We calculate the cross-entropy loss between the generated labels and the labels produced by the discriminator with the generator data as input. In this case, the generator loss is the loss encountered by the generator when discriminator classifies the adversaries generated by the generator as `attack' samples. The following equation says that we want to minimize the entropy for the adversarial examples generated by the generator, with respect to class `0', which corresponds to the `non-attack' class. This forces the generator to optimize in order to make sure that the discriminator gets confused and classifies the `adversaries' as `non-attack' samples.
        $min_G V_G(D, G)
         = \newline min_G (\mathbb{E}_{adversary \sim {}\\ 
        p_{adversary}}[log(D(G(adversary)))])$
    
    \item Discriminator Loss: This loss is calculated from the discriminator. It calculates how well the discriminator can distinguish between the fake data and the real data. The fake data or the generated data is given the label `1', and the real data is given the label `0'. We sample the real data and the fake data, in batches, as part of the training process and see the output of the discriminator. We calculate cross-entropy loss between the labels assigned to the data and the predictions of the model. In this case, the discriminator loss will be the usual cross entropy loss encountered by the classifier, and also the loss encountered by classifying the adversarial examples as `non-attack'.
    The following equation says that we want to minimize the entropy of the `non-attack' samples with respect to class `0' or the `non-attack' class. We also want to minimize the entropy of the `attack' samples and the adversarial samples with respect to class `1', or the `attack' class. This makes sure that the discriminator gets optimized in order to classify the `non-attack' samples as `non-attack', `attack' as `attack', and `adversaries' as `attack' respectively.

    $min_D V_D(D, G) = 
    \newline min_D (\mathbb{E}_{x_{non-attack} \sim p_{non-attack}}[log(D(x_{non-attack})] + \mathbb{E}_{attack \sim p_{attack}}[log(1-D(x_{attack}))] + \mathbb{E}_{adversary \sim p_{adversary}}[log(1-D(G(adversary)))])$

\end{itemize}

We use the generator to learn the data distribution in order to generate attack event samples. 

\subsection{Adversarial Attacks}
\label{advers_attack}
Adversarial Attacks fool a neural network classifier by making minuscule perturbations in the data samples in order to fool a neural network. This can be achieved in multiple ways. One of the ways this can be achieved is the `Fast Gradient Sign Method'.\cite{Goodfellow_fast_gradient} This method calculates the gradient of the loss function, when changing the label of the output to a desired label, with respect to the input data. This helps the attacker to calculate which direction they should change the data,  in order to force the neural network to produce the desired result as output. In this experiment, we calculate the gradient of the desired output (`non-attack') with respect to a sample which is drawn from the pool of `attack' data.

\begin{equation}
    \delta = \epsilon . sign(\nabla_{x_{attack}} loss(x_{attack}, y=`non-attack))
    \label{equation1}
\end{equation}

\begin{equation}
    x_{attack} = x_{attack} + \delta
    \label{equation2}
\end{equation}

In the equation \ref{equation1}, the `loss' refers to the cross-entropy loss of the trained neural network. The $\delta$ refers to the small perturbations which we use to change the data from the attack samples. If we disturb the input sample, with small perturbations for a finite number of iterations, we can reach a value for $x_{attack}$ which is numerically very close to the actual $x_{attack}$ sample, but forces the neural network to produce an output of `non-attack'. 

The Fast Gradient Sign method \cite{Goodfellow_fast_gradient} calculates the direction of partial derivatives of the data sample, with respect to the desired output. This results in a 26 dimensional variable, where each dimension says the direction in which we have to move in order to perturb the data effectively. We multiply the direction vector with a very small number $\epsilon$. This produces the `perturbation vector' which we add to the $x_{attack}$ sample. We do this for a finite number of iterations and check the output of the neural network, for this perturbed data. We also want to make sure that the perturbations are not affecting sensitive features. In Section \ref{Experimental Setup}, we will show how we have conducted experiments where we perform adversarial attacks without interfering with sensitive features.

\section{Experimental Setup and Results}
\label{Experimental Setup}
The GAN based classifier, has two neural networks. First the generator and then the discriminator. The generator is a 4 layer neural network with `4', `16', `16', and `26' neurons respectively. This generator results in a 26 dimensional vector which is a candidate for a forged `attack' sample. 
The discriminator is a 5 layer neural network, with `26', `16', `16', `2', and `1' neuron respectively. The discriminator produces a binary output, where `0' corresponds to a `non-attack' and `1' corresponds to an  `attack'. The generator generates `attack' samples and sends them to the discriminator hoping for the discriminator to mispredict them into the `non-attack' class. The discriminator receives the generated samples, and also data samples from the actual dataset, and tries to predict the `non-attack' samples as `0', and everything else as `1'. The loss function, we use to calculate the discriminator loss and the generator loss, is sigmoid cross-entropy. 

As stated in Section \ref{dataset_descr},the merged dataset has 6M samples. We split the dataset into 12 folds of 500k samples, and we calculated the scores for the `non-attack' samples and the `attack' samples. 

At each iteration, we generate 20 samples for the `attack' class from the generator. We sample 30 data points from the dataset to construct each batch. This brings the total number of data points for each batch being sent to the discriminator to 50. We sample in such a way, so that for each batch we have equal number of `attack' and `non-attack' samples. 

We run our experiments for the 500k samples, and then on the entire dataset of 6M samples. We split each dataset into training and testing with a ratio of 0.3. This means we have 350k samples for training, and 150k samples for test.  We report the scores precision, recall, and F-1 scores on the test set for the experiment. In Table \ref{precision table}, we can see the average scores of the test-sets of our 500k samples after 50,000 iterations. The `Non-Attack' scores signify the model's performance while detecting the samples from the test set, where the label is `0'. The `Attack' scores signify the model's performance while detecting the samples from the test set, where the label is `1' and also the generated samples from the generator. In Table \ref{precision table2}, we can see the performance of our model in the test set of the entire dataset.  

\begin{table}
\begin{center}
 \begin{tabular}{||c c c c||} 
 \hline
 Classes & Precision & Recall & F-1 Score \\ [0.5ex]
 \hline
 Non-Attack & 97 & 98 & 98 \\ 
 \hline
 Attack & 89 & 91 & 90 \\ [1ex]
 \hline
\end{tabular}
\caption{Precision, Recall, and F-1 score for our classes averaged across the splits.}
\label{precision table2}
\end{center}
\end{table}

\begin{table}
\begin{center}
 \begin{tabular}{||c c c c||} 
 \hline
 Classes & Precision & Recall & F-1 Score \\ [0.5ex]
 \hline
 Non-Attack & 98 & 96 & 97 \\ 
 \hline
 Attack & 85 & 92 & 88 \\ [1ex]
 \hline
\end{tabular}
\caption{Precision, Recall, and F-1 score for our classes for the entire dataset.}
\label{precision table}
\end{center}
\end{table}

In the second part of our experiment, we perform the adversarial attack as stated in the previous section. We perform `Fast-Sign Gradient Method' attack on the data samples. We sample 1000 data points from the dataset, making sure all the 1000 data points correspond to the `attack' label.The $\delta$ as mentioned in Equation \ref{equation2} is used to make small perturbations to each data point. We update the data points for 500 iterations, and then provide these changed values to the already trained classifier. We measure our attack success based on the ratio of the number of labels that have been changed, and the total number of samples sent to face the attack. The value of the $\epsilon$ stated in Equation \ref{equation1} is taken as 0.00005. This means that even after 500 iterations, the maximum change in a particular feature can face after perturbations is 0.0025, which is small considering our normalized features range from -1 to 1. 

As discussed in Section \ref{advers_attack}, we want to see how successful the attack is, if we do not want to interfere with sensitive components of the data. We calculate the partial derivatives of the loss function with the attack label, with respect to the input, as part of the process of the adversarial attack, as stated in Equation \ref{equation1}. Each partial derivative results in a 26 dimensional vector, signifying the direction, as well as the magnitude by which we have to move in order to reach a value for the sample which would produce the desired output. Although, in the `Fast-Sign Gradient Method'\cite{Goodfellow_fast_gradient} we are only interested in the direction or the `sign' of the gradients, we can also use the magnitude to calculate which features are significant for the adversarial attack to take place. We tabulate the values independently, and we mask the most significant features for the adversarial attack and re-calculate the attack success rates. The intuition behind this exercise is that if as a defence mechanism, we pay attention to the most significant features, we want to know if it is still possible to successfully attack the neural network based classifier with the help of the lesser significant features. We stop the $\delta$ updates, for those features and perform the updates for the rest of the features and calculate the attack success scores in Table \ref{adversarial attack table}. We can see, although the attack success rates drop significantly it is still possible to cause adversarial attacks on the remaining features with at least 41\% success rate. This number is still high, considering that its happening in a network intrusion detection system, where we want minimum room for error. 

\begin{table}
\begin{center}
 \begin{tabular}{||c c ||} 
 \hline
 Number of candidate features & Attack Success Rate \\ [0.5ex]
 \hline
 All present & 96 \\ 
 \hline
 Significant feature not used & 72 \\ [1ex]
 \hline
 Top 2 features not used & 60 \\ [1ex]
 \hline
 Top 3 features not used & 41 \\ [1ex]
 \hline
\end{tabular}
\caption{Adversarial Attack Performance based on the number of features used}
\label{adversarial attack table}
\end{center}
\end{table}

\section{Conclusion and Future Work}
\label{Conclusion}
In this paper, we have discussed how we can train a neural network based classifier for a network intrusion detection system, and can successfully break it using `Fast-Sign Gradient Method'. We show that adversarial attacks can cause harm even for network intrusion detection system. As opposed to domains such as computer vision where changing a few pixels, even drastically, may not appear to visually change the image, changing any dimension of the input vector may not be as easy in network intrusion detection systems. We show how even by not using sensitive features we are able to successfully perform adversarial attacks on the classifier. We have built a classifier and trained it with adversarial examples, to make it robust. But this particular method of adversarial attack is successful against breaking a neural network which has seen adversarial examples during training phase. In future, we would like to build defence mechanisms against these kind of attacks for network intrusion.

\bibliographystyle{plain}
\bibliography{references}

\end{document}